\documentclass[letterpaper, 10 pt, conference]{ieeeconf}  

\IEEEoverridecommandlockouts                              

\usepackage{algorithm}
\usepackage[noend]{algpseudocode}
\algrenewcommand\algorithmicrequire{\textbf{Input:}}
\algrenewcommand\algorithmicensure{\textbf{Output:}}
\usepackage{amsmath} 

\DeclareMathOperator{\atantwo}{atan2}
\usepackage{amssymb}  
\usepackage{graphicx}
\usepackage{siunitx}
\usepackage{svg}
\usepackage{tikz}
\usepackage{pgfplots}
\usepackage{layouts}
\usepackage{xprintlen}
\usepackage{atbegshi}
\usepackage{url}
\AtBeginShipoutFirst{\raisebox{1cm}{\parbox{0.9\textwidth}{\centering\normalsize Preprint of the paper which appeared in:\\ \large 2024 IEEE/RSJ International Conference on Intelligent Robots and Systems (IROS)}}}

\usepackage{todonotes}
\setuptodonotes{inline}

\title{\LARGE \bf
Efficient Dynamic LiDAR Odometry for Mobile Robots with Structured Point Clouds
}

\author{Jonathan Lichtenfeld, Kevin Daun and Oskar von Stryk$^{1}$
\thanks{All authors are with the Simulation, Systems Optimization and Robotics Group, Technical University of Darmstadt, Hochschulstr. 10, 64289 Darmstadt, Germany. {\tt\small\{lichtenfeld, daun, stryk\}@sim.tu-darmstadt.de}}%
\thanks{
Research presented in this paper has been supported in parts by the German Federal Ministry of Education and Research (BMBF) within the DRZ project (grant no. 13N16475), and the KIARA project (grant no. 13N16274), and by the LOEWE initiative (Hesse, Germany) within the emergenCITY center.
 }
}

\begin{document}
\maketitle
\thispagestyle{empty}
\pagestyle{empty}
\begin{abstract}
We propose a real-time dynamic LiDAR odometry pipeline for mobile robots in Urban Search and Rescue (USAR) scenarios. 
Existing approaches to dynamic object detection often rely on pretrained learned networks or computationally expensive volumetric maps.
To enhance efficiency on computationally limited robots, we reuse data between the odometry and detection module.
Utilizing a range image segmentation technique and a novel residual-based heuristic, our method distinguishes dynamic from static objects before integrating them into the point cloud map.
The approach demonstrates robust object tracking and improved map accuracy in environments with numerous dynamic objects.
Even highly non-rigid objects, such as running humans, are accurately detected at point level without prior downsampling of the point cloud and hence, without loss of information.
Evaluation on simulated and real-world data validates its computational efficiency.
Compared to a state-of-the-art volumetric method, our approach shows comparable detection performance at a fraction of the processing time, adding only 14 ms to the odometry module for dynamic object detection and tracking.
The implementation and a new real-world dataset are available as open-source for further research.
\end{abstract}

\section{INTRODUCTION}
The capability to localize itself and create a map of the environment is essential to enable autonomous navigation and other assistance functions with mobile robots in unknown disaster environments without GNSS reception.
The problem is referred to as Simultaneous localization and mapping (SLAM).
Many established approaches assume a static world, which is usually not the case in challenging Urban Search and Rescue (USAR) environments.
Often, there are rescue workers, vehicles, or other robots in proximity.
The need for dealing with such objects is motivated by three main aspects. 
Firstly, for autonomous navigation, obstacle avoidance has to be ensured to improve the semantic understanding of the environment and enable better autonomous functions, such as the consideration of dynamic objects in motion planning and control.
Secondly, dynamic objects need explicit handling for mapping as otherwise, they typically create "ghost trace" artifacts in the map and can potentially impede the quality of the localization in highly dynamic environments.
Thirdly, as a step towards human-machine-interaction, an understanding of the dynamic semantics in the environments enables novel approaches to human-robot interaction with the rescue teams or possible victims. 
For instance, a robot could follow a rescue worker walking in front of it or lead a potential victim to a safe area.

\begin{figure}
    \centering    
    \includegraphics[width=\linewidth]{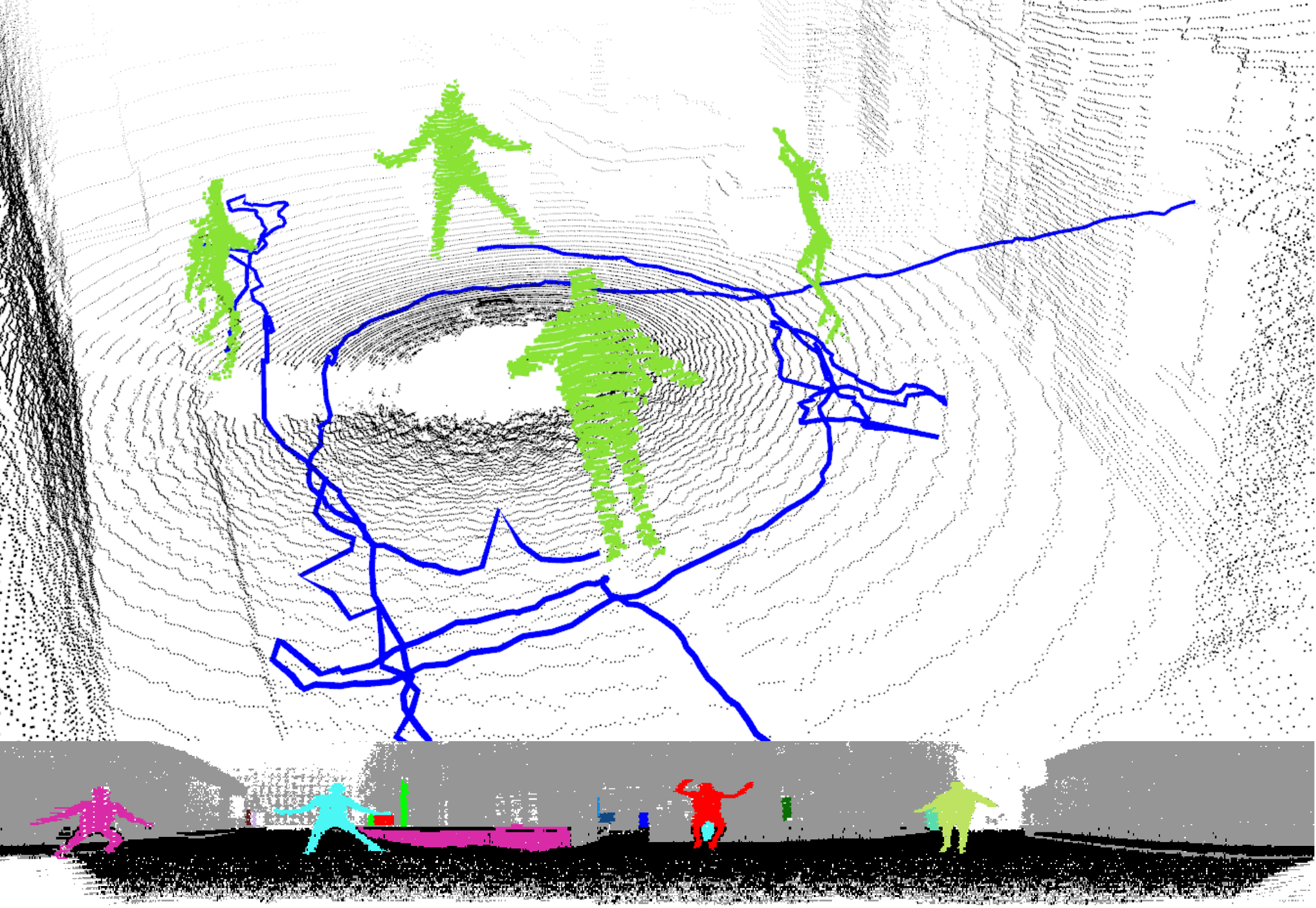}    
    \caption{Application of the proposed approach for dynamic LiDAR odometry, enabling the efficient segmentation and tracking of highly articulated objects.
        Top: A group of jumping persons is detected and tracked in a highly dynamic sequence. Dynamic points are shown in green, and trajectories are indicated in blue.
        Bottom: The range image used for object segmentation of the above cloud. Different colors represent different static and dynamic objects.
        }
    \label{fig:circle}
\end{figure}

Existing SLAM approaches considering the environment dynamics can provide very accurate environment segmentations but are typically either limited by computational expense, no online capability, or inaccurate representations of highly articulated objects.
For example, the post-processing method ERASOR~\cite{lim_erasor_2021} removes ghost traces from the map after the SLAM process, thus, no online handling of the objects is possible.
Learning-based methods that have become widely used in the context of autonomous driving \cite{yang_3dssd_2020, lang2019pointpillars, milioto_rangenet_2019}
allow the accurate online detection of (moving) objects but are computationally expensive and fail on instances of classes that are not included in the training data, rendering them unsuitable for USAR applications where potential objects are not necessarily known beforehand.

This work proposes a new approach to dynamic Light Detection and Ranging (LiDAR) odometry, which addresses the aforementioned limitations.
In contrast to other recent works, we follow a grid-free approach and operate directly on a structured point cloud, allowing for faster processing of high-resolution scans even in large-scale environments.
As a backbone, we use the state-of-the-art LiDAR odometry method proposed by~\cite{chen_direct_2022}.
Working on the range image, the point cloud is segmented, and objects are tracked individually so that they can be used in other processes, too.
Based on the observation that moving objects exhibit higher residuals in the scan matching process than static ones, we propose a novel heuristic to distinguish between the two classes.
Dynamic objects can then easily be removed from the scan before integrating it into the map.

Our approach is designed to consider the specific requirements~\cite{daun23} for USAR applications, which implies that its
mechanisms are as general as possible and rely only on few assumptions about the environment.
The main focus is on obtaining a light-weight, comprehensible
system that runs in real-time on ground-based robots using a single high-resolution LiDAR
sensor.
In summary, the main contributions of this work are:
\begin{itemize}
    \item A lightweight end-to-end pipeline for dynamic LiDAR odometry, including the components odometry, object detection, tracking, and mapping
    \item A novel residual-based heuristic to distinguish dynamic from static objects
    \item Evaluation on a public benchmark and created real-world data
    \item The open-source release of the code and dataset\footnote{\url{https://github.com/tu-darmstadt-ros-pkg/dynamic_direct_lidar_odometry}}.
\end{itemize}    

\section{RELATED WORK}
Most LiDAR Odometry and Mapping methods treat dynamic objects as outliers in the scan matching process \cite{zhang_loam_2014,shan_lego-loam_2018, chen_direct_2022}, which tends to worsen the odometry estimate and thus the map quality.
A complementary aspect to this problem is the Detection and Tracking of Moving Objects (DATMO).
Although the majority of publications cover only either of them \cite{chen_direct_2022, schmid_dynablox_2023, zhang_loam_2014, shan_lego-loam_2018}, DATMO can be integrated into the SLAM process, leading to higher fidelity maps and more accurate pose estimates in dynamic environments.
At the same time, a precise map helps in providing self-localization and detecting objects. 
Wang \textit{et al.}~\cite{wang_online_2003} were the first to propose a framework that integrates both problems in a complementary manner by detecting moving objects in each scan before obtaining the pose update from a SLAM module. 
Since tracking of dynamic objects is well studied in the computer vision community and robust off-the-shelf multi-object trackers exist, a key challenge of DATMO lies in the detection of moving objects in the first place.

Following the classification by Yoon \textit{et al.}~\cite{yoon_mapless_2019}, moving object detection techniques can be categorized into three groups: model-based detectors, methods that utilize a priori known maps, and approaches that rely on the most recent data only. 
Model-based, or class-specific, methods can provide accurate detections for predefined object classes in both camera images \cite{bolya_yolact_2019} and point clouds \cite{milioto_rangenet_2019,yang_3dssd_2020}.
Considering only frame-wise detections, one cannot tell which objects are moving.
Semantic labels, however, can be used to distinguish possibly moving and most likely static objects.
For example, vehicles, animals, and pedestrians belong to the former category, while buildings, traffic lights, and trees belong to the latter. 
Nevertheless, purely model-based classifiers fail to differ between temporarily static and actually moving objects, such as parked and driving cars.
Especially in USAR environments, objects of many different classes are present, which can hardly all be represented in training data.

In some use cases for mobile robots, a map of the static environment already exists. This applies, for instance, to autonomous forklifts deployed in warehouse logistics or surveillance robots in buildings. If a model of the building is given or the robot has captured a motion-free map before, it is relatively easy to detect discrepancies in the robot's perception and the stored data using change detection techniques \cite{xiao_street_2015}.
These discrepancies can indicate moving objects.

Map-free methods can only rely on the most recent sensor data. 
A way to detect changes without a given map is the online construction of occupancy grids.
Typically, the space is discretized into voxels and each voxel is assigned a probability of occupancy based on ray-tracing operations~\cite{azim_detection_2012}.
Postica \textit{et al.}~\cite{postica_robust_2016} then derive moving objects from inconsistencies between free space and occupied space.
The work by Schmid \textit{et al.}~\cite{schmid_dynablox_2023} is based on a similar principle, but builds on the Truncated Signed Distance Fields (TSDF) framework Voxblox~\cite{oleynikova_voxblox_2017} for a more efficient map representation.
This approach requires highly accurate localization information, which in turn is affected by the presence of moving objects.
\cite{schmid2024khronos} considers both short-term and long-term dynamics by constructing a novel dense spatio-temporal representation of the robot environment.
However, it focuses on indoor RGB-D applications, incorporating also semantic information that is difficult to obtain with LiDAR sensors.
Dewan \textit{et al.}~\cite{dewan_motion-based_2016} use motion models derived from subsequent scans for dynamic detection. 
Employing Random Sample Consensus (RANSAC), they estimate motion models for the static scene and dynamic objects. 
Subsequently, each point is assigned to a motion model following a Bayesian approach. 
Their evaluation on real world scenes shows high detection and tracking accuracy. 
However, the proposed motion models are incompatible with non-rigid objects, resulting in a failure to detect articulated objects such as humans or animals.
Besides, no information on the used hardware or processing times is given, which also applies to a similar work by Moosmann \textit{et al.}~\cite{moosmann_joint_2013}. 
A follow-up work by the same authors computes dense scene flow improved by incorporating a deep neural network~\cite{dewan_rigid_2016}.
Wang \textit{et al.}~\cite{wang_drr-lio_2023} propose a vertical voxel occupancy descriptor to discriminate static and dynamic regions.
Underwood \textit{et al.}~\cite{underwood_explicit_2013} compare two point clouds and include free space reasoning of unmeasured areas by incorporating ray tracing. 
This procedure is facilitated by using a spherical coordinate space for the point representation~\cite{ferri_dynamic_2015,pomerleau_long-term_2014}.
Dynamic detection across a multi-session map has been studied in~\cite{ding_multi-session_2018}, also using ray casting and a voxelized point cloud represented as a kd-tree.

Lastly, there are methods that remove dynamic points during post-processing~\cite{lim_erasor_2021,schauer_peopleremoverremoving_2018}, using not only past knowledge but also future scans.
Typically, the processing time per scan far exceeds the sensor's frame rate.
This renders them unsuitable for real-time applications, despite their capability to build highly accurate static maps.

The work by Mao \textit{et al.}~\cite{mao_lio-dor_2023} is the most similar to ours, since it also comprises a combined odometry and detection approach. 
The main difference is that they apply a spatial region growing algorithm for clustering points.
This increases the computation time for high-resolution point clouds if no down-sampling is applied.
Dynamic object detection is then based on the comparison of the bounding box overlap between the current and the previous frame.
We assume that this approach struggles with slow-moving objects, as the bounding box overlap is too high to detect them, or with ambiguous associations between scans with multiple objects close to each other.
We address this issue by tracking objects over longer periods.

This work uses new scans and, to a small extent, the online constructed map to detect moving objects in previously unknown surroundings.
As one of few works only, we exploit the data structure of modern LiDAR sensors for speeding up the object detection.
Some weak assumptions about the model's shapes are made, hence it is not completely model-free.
Instead of a dense map representation \cite{schmid_dynablox_2023, schmid2024khronos}, we choose raw point clouds to maintain efficiency even for large-scale scenes.

\begin{figure}[tpb]
  \centering
  \includegraphics[width=\columnwidth]{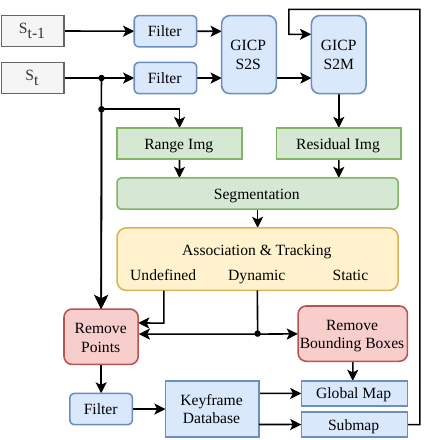}
  \caption[System overview]
  {
    System overview. Two consecutive scans are registered using the odometry module (blue). The detection module (green) segments the projected range image and residual image into individual objects. The Tracking and Association module (yellow) assigns a dynamic state to each object.
    Non-static points are removed from the current scan before integrating it into the keyframe database. Ghost traces are removed from the global map (red) based on the objects' bounding boxes.
  }
  \label{fig:overview}
\end{figure}

\section{METHOD}
We propose a novel approach to dynamic LiDAR odometry with a focus on identifying and tracking distinct dynamic objects.
The concept can be divided into an odometry, detection, tracking and point removal modules that influence one another. 
A complete overview is given in Fig.~\ref*{fig:overview}.

\subsection{Overview}
Input to the pipeline is a series of laser range scans.
We build upon and extend the LiDAR odometry method DLO proposed by Chen \textit{et al.}~\cite{chen_direct_2022} which employs a two-stage Generalized Iterative Closest Points (GICP) approach for fast registration.
The first stage computes an initial guess of the robot's motion by aligning two consecutive scans $\mathcal{S}_{t-1}$ and $\mathcal{S}_{t}$.
The obtained transformation is then used as a prior for the second stage, which aligns $\mathcal{S}_{t}$ to a submap $\mathcal{M}_{t}$ for fine registration.
Both source and target cloud are stored as kd-trees for efficient nearest neighbor searches.
For more details, we refer the reader to the original work.
The remaining residuals after convergence are projected onto an image, which in turn is combined with the range image to segment the input into individual geometric objects.
The Tracking and Association module then updates a Kalman filter for each object, associates new detections to existing objects, and assigns a dynamic state to each object.
Finally, we remove those points in the current scan referring to segments labeled as non-static before integrating the scan into the keyframe database.
The latter is used to generate submaps and the global map.

\subsection{Detection}
To detect dynamic elements in the scene, we first detect relevant segments as objects by performing clustering with a range image representations of the scan cloud. Then, we utilize the scan matching residuals to classify the segments into dynamic and static.

\subsubsection*{Range Image-based Segmantation}
Geometric segmentation of point clouds is expensive because spatial clustering requires nearest neighbor searches, which is costly for large point clouds.
Instead, we employ the established range image segmentation by Bogoslavskyi \textit{et al.}~\cite{bogoslavskyi_fast_2016}.
First, the point cloud $P$ is projected onto a cylindrical image $I$ by a mapping $\Pi \colon \mathbb{R}^3 \to \mathbb{R}^2$. 
Each point $\textbf{p}_i=(x,y,z)^\text{T} \in P$ is assigned to an image coordinate $(u,v)$ via
\begin{equation}
\left(
\begin{array}{c}
u\\v
\end{array}
\right)
=
\left(
\begin{array}{c}
  \Bigl\lfloor\left[f_{up}-\arctan\left(\frac{z}{\sqrt{x^2+y^2}}\right)\right] \alpha_v^{-1}\Bigr\rfloor \\[4pt]
  \lfloor\atantwo(x,y)\,\alpha_h^{-1}\rfloor
\end{array}  
\right)
\enspace ,
\label{eqn:image_projection1}
\end{equation}
where $f_{up}$ denotes the upper aperture angle, $\alpha_h$ the horizontal and $\alpha_v$ the vertical sensor resolution, respectively. 

This approach can be applied to any point cloud, regardless of their sparsity and point order. 
The image pixels store the range, i.e. the distance of the regarding point to the sensor
\begin{equation}
    I(u(x,y,z),v(x,y,z))=\sqrt{x^2+y^2+z^2} \enspace.
\end{equation}
Pixels without a belonging point are marked as invalid. 
If the point cloud is structured with height $h$ and width $w$, each point $\textbf{p}_i$ directly represents a pixel. Assuming a row-major layout, the mapping simplifies to
\begin{equation}
\left(
\begin{array}{c}
u\\v
\end{array}
\right)
=
\left(
\begin{array}{c}
 \lfloor\frac{i}{w}\rfloor\\
  i \mod w
\end{array}  
\right)
\enspace .
\label{eqn:image_projection2}
\end{equation}

Prior to the object segmentation, we apply a coarse ground removal to the range image as proposed in \cite{shan_lego-loam_2018}. 
The remaining points are clustered into segments as described in \cite{bogoslavskyi_fast_2016}.
Starting from the upper left point in the range image with label $1$, all direct neighbors of that point are considered candidates for this segment. 
Since the range image is a cylindrical projection, the left and right image edges are wrapped.
If a candidate fulfills a heuristic condition and is not yet labeled as part of a segment, ground, or invalid, it is added to a queue and receives the label of the current segment. 
All points in that queue are treated this way, iteratively, until the queue is empty. 
In this case, the label is increased by $1$ and the process is repeated, starting with the next unlabeled point.
More information about the chosen heuristic can be found in \cite{bogoslavskyi_fast_2016}.

\subsubsection*{Scan-Matching Residual-based Classification}
\begin{figure}[tp] 
  \centering
  \smallskip
  \includegraphics[width=\columnwidth]{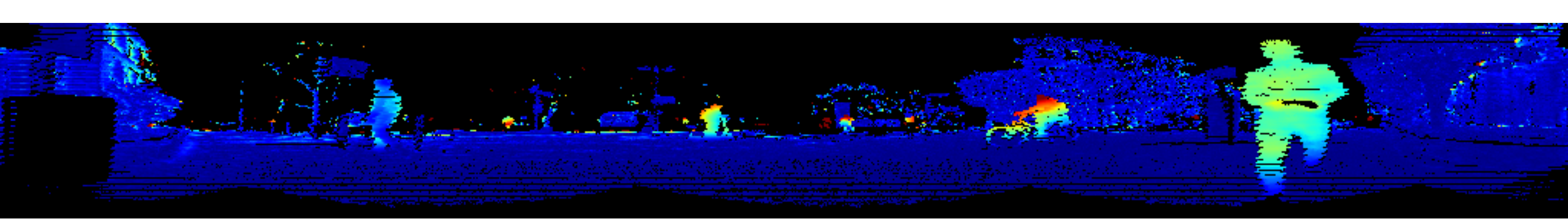}
  \caption
  {
    Residual image as obtained from the GICP algorithm.     
    For better visibility, the registration was performed on the original point cloud, whereas the method uses the downsampled and filtered cloud.
    A rainbow color map highlights the points with higher residuals.
    Black pixels represent invalid points.
  }
  \label{fig:residual_img}
\end{figure}
\begin{algorithm}[tp]
\centering
    \caption{Average Residuals Computation}
    \label{alg:residuals}
    \begin{algorithmic}[1]
    \Require{Segment $S$, residual image $R_{ICP}$}
    \Ensure{Average of residuals $r_{avg}$}
        \State $n \gets 0, r \gets 0$
        \For{$(u,v)\, in\, S$}
          \If{$R_{ICP}(u,v) \neq 0$}
            \State $r \gets r + R_{ICP}(u,v)$
            \State $n \gets n + 1$
          \EndIf
        \EndFor
        \State \textbf{return} $r/n$ \textbf{if} $n > 0$ \textbf{else} $0$
    \end{algorithmic}
\end{algorithm}
The above segmentation technique usually yields many segments, depending on the complexity of the environment and chosen parameters.
To classify between static and dynamic objects at the tracking stage, we propose the use of so-called residual images.
They are built from the residuals of the GICP algorithm implemented in \cite{chen_direct_2022}.
It iteratively searches for nearest neighbor correspondences between two point clouds and finds a transformation $\mathbf{T}$ that minimizes a cost function such that
\begin{equation}
  \mathbf{T}=\underset{\mathbf{T}}{\arg \min } \sum_i d_i^{(\mathbf{T})^{T}}\left(\mathcal{C}_{i}^{\mathrm{t}} + \mathbf{T} \mathcal{C}_{i}^{\mathrm{s}} \mathbf{T}^{T}\right)^{-1} d_i^{(\mathbf{T})}
\label{eqn:gicp}
\end{equation}
with the covariance matrices of the target and source point $\mathcal{C}_{i}^{\mathrm{t}}$ and $\mathcal{C}_{k}^{\mathrm{s}}$, respectively, and the distance $d_i^{(\mathbf{T})} = p^t_i - \mathbf{T}p^s_i$ between the corresponding points.
After convergence, a residuum remains for each point in the source cloud, which expresses the Euclidean distance to its nearest neighbor in the target cloud.
Considering the scan-to-map module, the source is the current scan, and the target is the current submap. 
The idea in the following is that moving objects are not represented in the submap, which serves as the target cloud. 
Therefore, points belonging to these objects tend to have higher residuals compared to the rest of the points, as their distance to the nearest neighbor is greater.
We project the point-wise residuals onto an image $R_{ICP}$ (see Fig. \ref{fig:residual_img}) using Equation \eqref{eqn:image_projection1}.
Due to the sparsity of the preprocessed point cloud used for registration, most pixels remain empty.
The average residuals for each segment can be computed by averaging the pixel values within the segment's mask, as shown in Algorithm \ref{alg:residuals}. 
These average residuals are stored for each segment and are used in the dynamic state update step of the tracking pipeline, which will be explained in the next section. 
One key advantage of these residual images is that they are a byproduct of the odometry module and require minimal additional computation, making them easy to integrate into the pipeline.

\subsection{Tracking and State Update}
\subsubsection{Tracking and Association}
It is crucial that objects are tracked in the global frame $S_W$, because movements parallel to the robot appear static in the robot’s frame $S_R$.
At each time step, the detection module provides a set of detections.
Since each image pixel refers to a unique point in the cloud, we can directly recover their 3D shapes from the point cloud.
For an effective and compact state representation, we compute the object's bounding box through Principal Component Analysis (PCA) and store the point indices as well as the average residuum of the segment.
A Kalman filter is then initialized for each object, which internally uses a 10-dimensional state representation consisting of the object's position, rotation around the $z$-axis, bounding box dimensions and velocity.
At each time step, the filters are updated with the associated detections.
Upon receiving new detections, we associate them with existing objects, using the approach proposed by Weng \textit{et al.}~\cite{weng_ab3dmot_2020} which consists of Kalman filter updates, data association and a Birth-Death memory.
Data association is based on the Hungarian algorithm \cite{kuhn_hungarian_1955} which minimizes the sum of the costs of the associations.
We assign the cost $c_{i,j}$ between the $i$th tracked object and $j$th detection 
\begin{equation}
    c_{i,j} =\alpha\cdot (1-d_1(BB_i,BB_j)) + \beta \cdot d_2(|p_i|,|p_j|)
\end{equation}
where $\alpha$ and $\beta$ are weights, 
$d_1$ is the Intersection over Union (IoU) between the bounding boxes
and $d_2 = 1 - \frac{\min(\lvert p_i\rvert, \lvert p_j\rvert)}{\max(\lvert p_i\rvert, \lvert p_j\rvert)}$
relates the number of points in both objects.
Then, we update each object's Kalman filter with the associated detection.
Objects without matches for $n$ consecutive steps are removed.

\subsubsection{Dynamic State Update}
\begin{figure*}[tpb]
  \centering
  \smallskip
  \includegraphics[width=.95\textwidth]{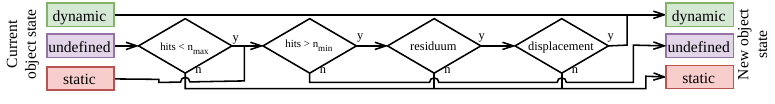}
  \caption[Dynamic state update]
  {
  Dynamic state update.
    At each time step, all tracked objects are updated and can change their dynamic state based on the number of detections (hits), their average residuum and the displacement from their origin.
  }
  \label{fig:dynamic_state_update}
\end{figure*}
After the data association step, we update the state of each object (see Fig. \ref*{fig:dynamic_state_update}).
An object can be in one of three states: \textit{undefined}, \textit{static}, or \textit{dynamic}. The default state for new objects is \textit{undefined}.
If it does not satisfy the conditions for a state transition to \textit{dynamic} within $n_{max}$ detections, we mark it as \textit{static}. 
To transition from \textit{static} or \textit{undefined} to \textit{dynamic}, we check the following conditions:
First, a minimum number of $n_{min}$ detections is required to avoid false positives.
Then, the average residual of the object's segment in the residual image $R_{ICP}$ is considered.
We observe that the average residual of an object tends to increase with its height, as the closest corresponding points are often ground points. Therefore, using a static threshold would either exclude all small objects if set too low or include almost all objects if set too high.
Instead, we check for each object if it fulfills the condition
\begin{equation}
r_{avg} \geq \theta_{res} h_S
\label{eqn:residual_condition}
\end{equation}
where $h_S$ is the range of z-coordinates belonging to the segment and $\theta_{res}$ is a heuristic threshold.
Lastly, we check if the object has moved by at least $\theta_{disp}$ meters from the start position where it was detected for the first time.
If all these conditions are met, the object is marked as \textit{dynamic} and keeps this attribute for its lifetime. This is to reliably filter objects out of the map that are not permanently moving, e.g. a car that stops at a traffic light for several time steps.
Another state, e.g. \textit{temporarily static}, could be introduced to handle such cases but is not considered in this work.

\subsection{Dynamic Points Removal}
The tracking module forwards the indices of all points that belong to dynamic objects to the odometry module.
These points can directly be removed from the transformed input scan. 
Afterward, the filtered and downsampled cloud is added to the keyframe database from which the submaps are generated.
To prevent ghost traces in the global map caused by initially static objects, we maintain a rolling window history of the bounding boxes of all objects.
If a static object turns dynamic, we remove all points from the global map that are within these bounding boxes.
As the global map is used for visualization only, this step is not time-critical for the odometry computation and is therefore  performed asynchronously.
Note that dynamic points are removed \textit{after} the scan-to-scan registration step, which means that they are present in the GICP module's source cloud but not in the target cloud. 
This is due to the fact that the segmentation step is performed on the input cloud after transforming it into the global frame. 
The required transform is obtained as the result of the registration, which leads to a "chicken-egg-problem".
One could use the registration result as a first guess, then perform the detection and point removal on the transformed cloud, and finally repeat the registration with the cleaned input cloud. 
However, this would require a second costly kd-tree generation for the new cloud.
The presented procedure shows promising results despite this inaccuracy.
\section{EVALUATION}
The evaluation is performed on real-world data and a simulated public benchmark and results are compared to Dynablox~\cite{schmid_dynablox_2023}.
We focus on the accuracy of both detection and tracking of moving objects.
Since close- and mid-range objects are most relevant for rescue robots, we highlight them in the results. 
Furthermore, the influence of moving objects on the constructed map and the processing time is investigated.

\subsection{Experimental Setup}
We use a four-wheeled differential drive robot for the data recording, which is equipped with an Ouster OS-0 128 LiDAR sensor mounted on top for an occlusion-free 360\textdegree{} view. It provides structured point clouds of size $128 \times 1024$ at a rate of 10 Hz.
The experiments are conducted on a consumer-grade laptop with an Intel Core i7-10875H CPU and 32 GB of RAM and are therefore representative for applicability on mobile robots.

\subsection{Datasets}
To the author's best knowledge, only few public benchmark datasets with structured undistorted point clouds are available, of which very few were recorded on mobile ground robots in typical robotic environments.
Therefore, we recorded our own dataset \textit{kantplatz}, which is publicly available\footnote{https://tudatalib.ulb.tu-darmstadt.de/handle/tudatalib/4303}.
The entire trajectory is about \SI{500}{\meter} long and has a duration of \SI{480}{\second}.
It is recorded in an urban environment with a variety of moving objects, such as pedestrians, cyclists, cars, and buses.
Two persons are walking close to the robot during the entire recording, which serves for the evaluation of the detection and tracking performance.
To compare on public benchmark data, we also evaluate our method on the \textit{small town simulation} sequence of the DOALS dataset \cite{pfreundschuh_dynamic_2021}.
It exhibits simple geometric shapes and more complex rigid objects such as animals that move along predefined looped trajectories.
The point cloud structure is $64\times2048$. We changed the original layout from column-major to row-major format to meet the requirements of our method. 

\subsection{Point Classification Accuracy}
For better comparability, we limit the detection range of both approaches to \SI{25}{\meter}, i.e.
farther objects are not considered and only points within that range are included in the evaluation.
However, a direct comparison is made difficult by the large number of parameters in both approaches. We have adjusted them manually with reasonable effort to achieve optimal results.

\begin{figure}[tpb]
    \centering
    \smallskip
    \includegraphics[width=\columnwidth]{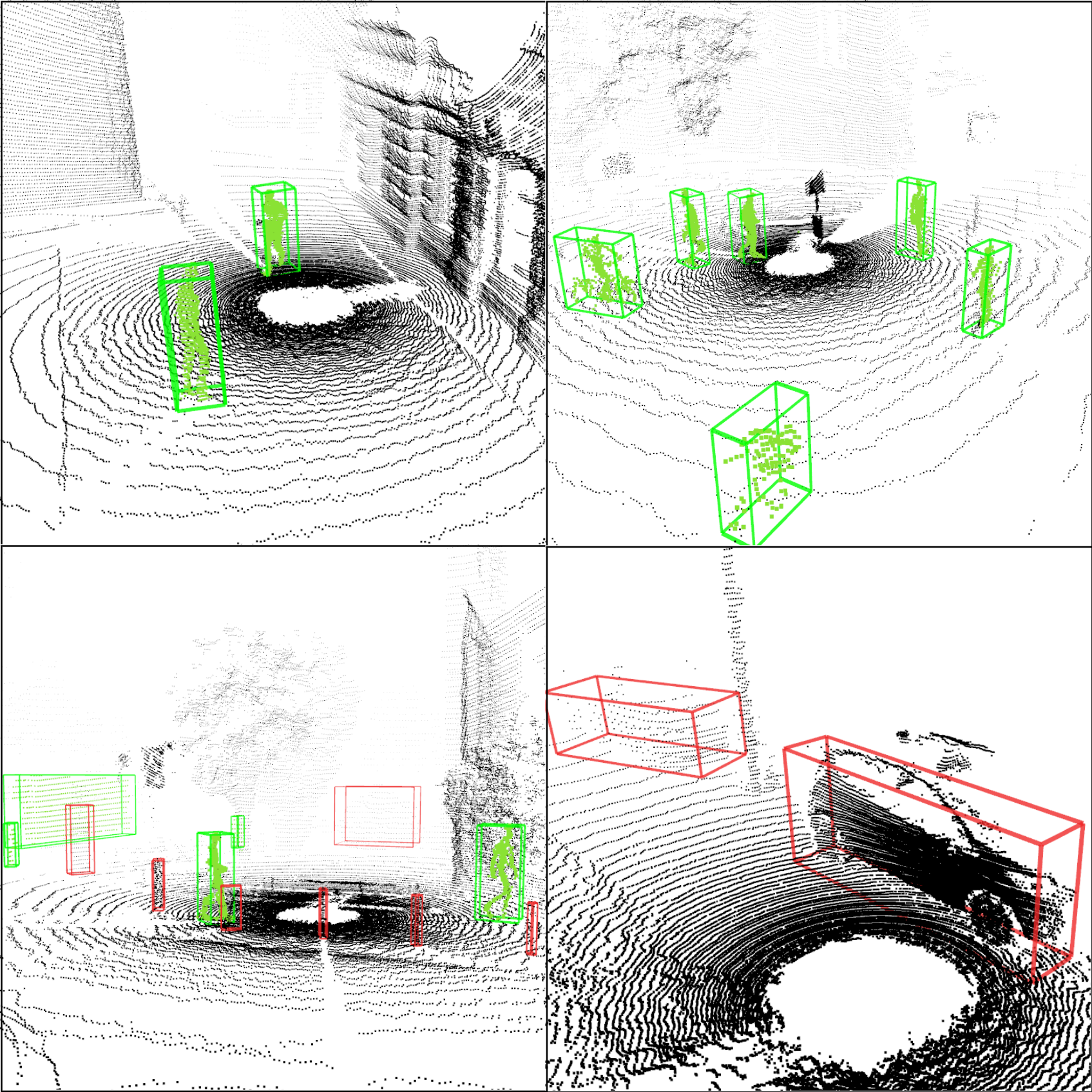}
    \caption{
        Segmentation examples of the \textit{kantplatz} dataset. Dynamic objects indicated by green bounding boxes and points, static objects by red bounding boxes.
        Top: Several pedestrians and cyclists are detected. 
        Bottom left: Both dynamic and static objects, such as street posts. A passing van on the left is detected, another car in the background is wrongly classified as static.
        Bottom right: Parked cars detected as static.
        }
    \label{fig:segmentation_examples}
\end{figure}

\begin{figure*}[tpb]
    \centering
    \smallskip
    \includegraphics[width=\textwidth]{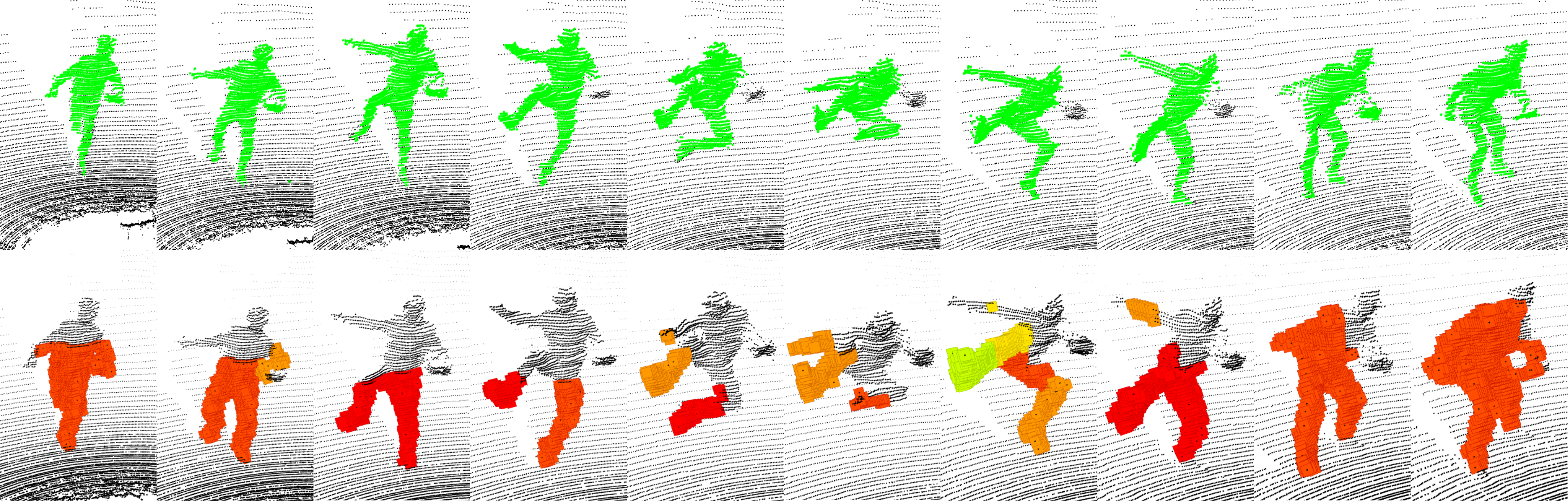}
        \caption{
        Dynamic scene from the \textit{kantplatz} dataset. Our approach (top) accurately detects the jumping person, only one arm is partially cut off. Dynablox (bottom) detects only parts of the person and tends to over-segment it.
        }
    \label{fig:jump}
\end{figure*}

On the real-world data, our method is able to detect and track most of the moving objects, as shown qualitatively in Fig. \ref*{fig:segmentation_examples}.
Even fast, non-rigid motions are detected with high accuracy (Fig. \ref{fig:jump}) whereas Dynablox often detected only segments of the objects.
Newly detected dynamic objects transition from \textit{undefined} to \textit{dynamic} after five time steps on average, depending on their velocity and the chosen thresholds. 
This is a good compromise between the detection of slow moving objects and the filtering of false positives such as static objects or noisy sensor readings.
Storing the history of static objects is particularly useful for the detection of slow moving objects, as they are detected as \textit{dynamic} only after they have moved a certain distance. 
If they have been labeled as \textit{undefined} in the previous scans, the detection module removes them from the scan before integrating it into the map.
If they have been marked as \textit{static}, the history is used to reliably remove them from the map after they have been detected as dynamic. 
This is particularly important for the beginnings of the recordings, where some objects are not yet moving, but it can also lead to the removal of few false positive points inside the respective bounding boxes.

\begin{figure}[tbp]
    \centering
    \input{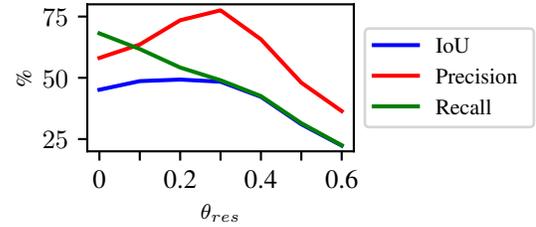}
    \vspace{-10pt}
    \caption{IoU, precision, and recall on the DOALS \textit{small town simulation} sequence for different values of the residuum threshold (see Eq. \ref{eqn:residual_condition}).}
    \label{fig:res_threshold}
\end{figure}

We quantitatively compare the detected dynamic points from the \textit{small town simulation} sequence to the ground truth annotations.
The average IoU, precision, and recall (\num{0.48}/\num{0.78}/\num{0.49}) stay behind the results of Dynablox (\num{0.69}/\num{0.99}/\num{0.69}).
We explain this mainly by the fact the dynamic objects in the sequence move in repetitive patterns, i.e. they cross the same area multiple times.
If a distant object is not detected due to its sparse point cloud representation, it can lead to traces in the submap.
These traces considerably reduce the residuals of objects at that location at a later time, which in turn prevents them from being detected as dynamic.
In contrast, if we omit the residuum check and base the dynamic detection solely on the distance that the object has moved, we risk FP dynamic detections. 
This is because depending on the angle from which the object is observed, its bounding box can vary significantly between two scans, thus leading to an apparent movement.
Fig. \ref*{fig:res_threshold} shows the trade-off between FP and FN point detections for different residuum thresholds.
Unsurprisingly, a volumetric mapping approach like Dynablox is more robust to these issues, as it can rely on the entire submap to classify each point.

\begin{figure}[tpb]
    \centering
    \includegraphics[width=\columnwidth]{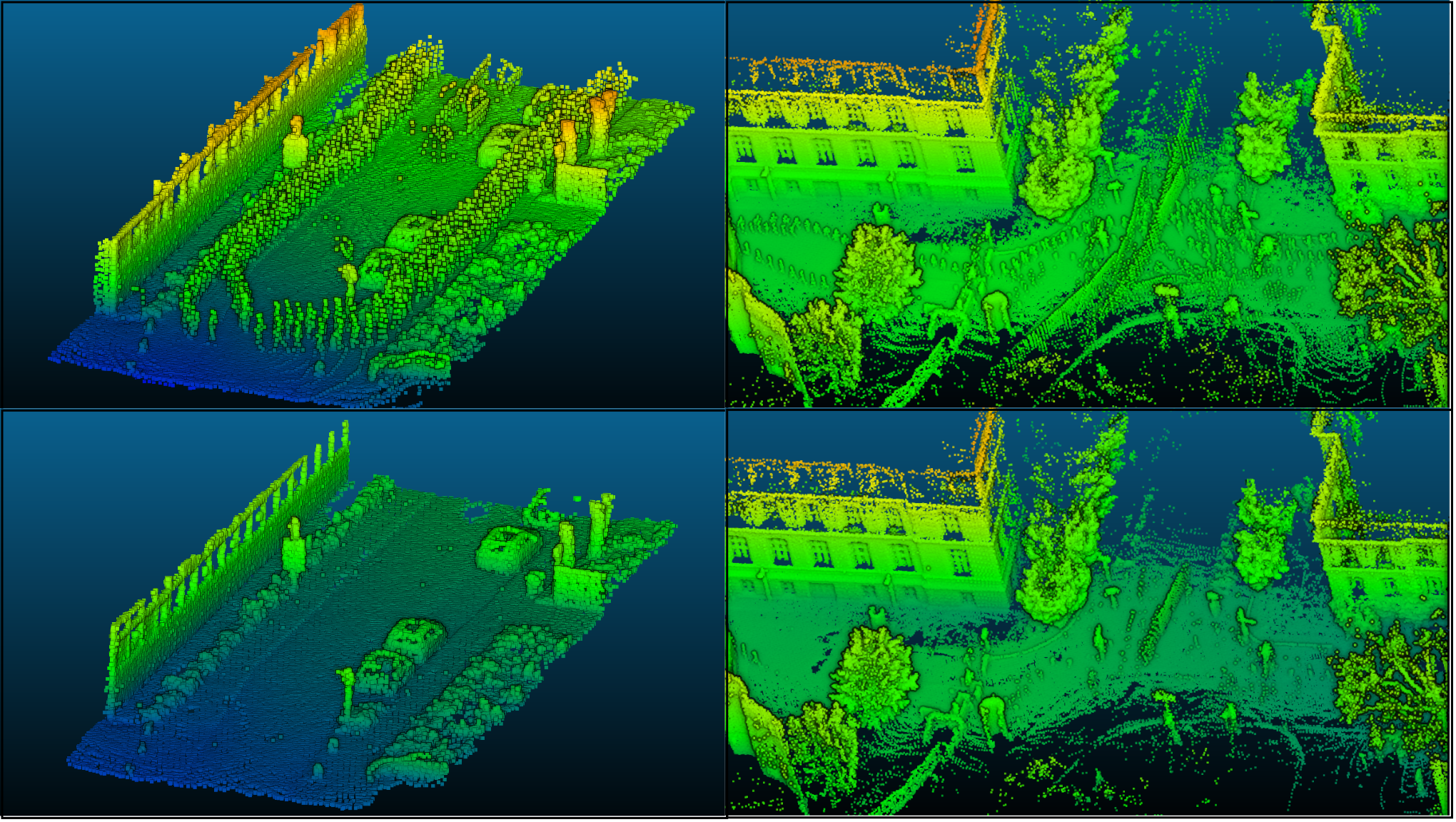}
    \caption{
        Comparison of the mapping results on the \textit{kantplatz} dataset. Top: Pure DLO~\cite{chen_direct_2022} without dynamic object handling, bottom: our approach.
        }
    \label{fig:map}
\end{figure}

\subsection{Tracking and Map Quality}
Evaluation of the \textit{kantplatz} sequence shows that the tracking performance is very robust.
The two persons following the robot are tracked throughout the entire sequence of about \SI{480}{s} duration.
For one of them, the trajectory is interrupted three times, i.e. the ID changes. This occurred twice when the person was occluded for several scans, and once when both persons were standing close to each other and appeared as one segment in the range image.
Fast and abrupt movements, such as jumping or running, are tracked well.
As a stress test, a group of four persons running in circles around the robot is tracked correctly (Fig. \ref*{fig:circle}).
Consequently, the map quality is improved by the accurate tracking (Fig. \ref*{fig:map}).
Some ghost traces are still visible in the map that mostly arise from remote objects. This could be tackled by tuning the parameters of the image segmentation.
In terms of localization accuracy, we did not find any significant differences compared to pure DLO. Despite the large number of moving objects, the ratio of dynamic to static points is too low to have a strong influence on localization. This aspect could be specifically investigated in a follow-up study.

\subsection{Processing Time Analysis}
\begin{figure}[ptb]
    \centering
    \input{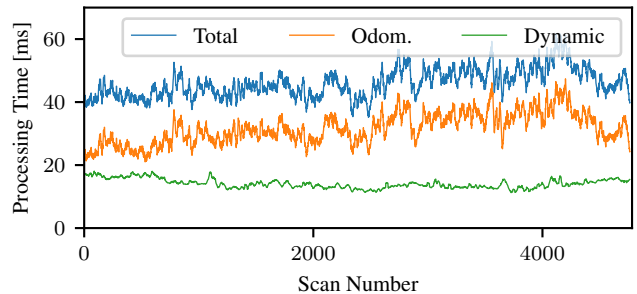}
    \caption{
        Processing time analysis of the \textit{kantplatz} sequence.
        Smoothed with a filter kernel of size 10 for better readability.}
    \label{fig:runtime}
\end{figure}

As one of our main motivations is to provide a lightweight solution, we analyze the processing time of our method by its algorithmic components (Table \ref*{table:runtime}).
On average, the pipeline takes \SI{46.0}{\ms} to process one scan from the \textit{kantplatz} sequence, which equals a frame rate of 21.7 Hz.
We emphasize that this includes the computation of the odometry, which is the most time-consuming part and depends strongly on the down-sampling of the point cloud.
The initial filtering reduces the number of points to $5\%$ of the original size, which is sufficient for the scan matching process.
The overhead of the dynamic object detection is only \SI{14.3}{\ms} on average.
As Fig. \ref*{fig:runtime} shows, the dynamic detection time is almost constant, while the odometry time increases slightly with the number of scans.
Looking at the individual components, we notice that the tracking time depends on the number of tracked objects, but is negligible.
Due to the structure of the point clouds, the projection time is also small.
The dense volumetric mapping approach used by Dynablox requires significantly more processing time and did not run in real time in our experiments.
It should be noted, though, that the dense voxel map created by Dynablox, which contributes to the high computation cost, provides more valuable information regarding scene understanding than our raw point cloud map.

\begin{table}[ptb]
    \centering
    \caption{Processing time [ms] for a single scan from an Ouster LiDAR sensor}
    
\begin{tabular}{l|cc}
    & kantplatz (OS-0 128) & small town simulation (OS-1 64) \\
   \hline
    odometry & $31.7 \pm 12.7$& $31.7 \pm 11.4$\\
    projection & $4.2 \pm 0.8$ & $3.5 \pm 0.7$ \\
    segm. & $9.0 \pm 1.3$& $4.4 \pm 0.8$ \\
    tracking & $1.1 \pm 0.6$ & $0.3 \pm 0.3$\\
    \hline
    \textbf{total} & $46.0 \pm 12.9$& $40.0 \pm 11.5$\\
    \hline
    \textbf{Dynablox} & $138.2\pm27.0$ & $111.0 \pm 12.2$\\
\end{tabular}
    
    \label{table:runtime}
\end{table}

\subsection{Limitations}
The most frequent failure case is the detection of false positives. As explained before, they can arise from the changing bounding boxes of static objects viewed from different angles. A more robust dynamic threshold could improve this, but carries the risk of delayed detection of actually moving objects.
Another source of false positives are apparent motions caused by the robot's own motion: If the robot passes by an opening, the captured segment of a wall behind it seems to move opposite to the robot's motion. 
This special case is handled better by occupancy grid-based methods.
In areas that are crossed by objects multiple times, the detection of dynamic objects is less reliable. Once an undetected object leaves a trace in the submap, it deteriorates the residuals of other objects at that location, which in turn prevents them from being detected as dynamic.

The method works best for structured point clouds that provide dense range images. 
In theory, unstructured point clouds could be used as input as well, but tests show that the range image projection and segmentation suffer in this case because of inaccurate point-to-pixel assignments. Furthermore, the processing time increases due to the additional computation of the projection.

\section{CONCLUSION}
In this work, we presented a novel approach to combine LiDAR odometry with light-weight dynamic object detection and tracking.
In contrast to other methods, our method does not rely on any pre-trained data and avoids the computational burden of volumetric mapping approaches by detecting dynamic objects directly in the structured point cloud.
Efficiency is gained by reusing residuum information from the scan matching process in combination with  heuristics to distinguish between dynamic and static objects.
We qualitatively evaluated our method on a public dataset and a new custom dataset, demonstrating significant improvements in the computational efficiency in comparison to a state-of-the-art volumetric mapping approach.
We demonstrated that our approach is able to detect articulated and fast-moving objects in mid-range distances and track them over long sequences.
Both the data and the code are available as open-source.

\addtolength{\textheight}{-12cm}   

\bibliography{bibliography}


\end{document}